\documentclass[journal]{IEEEtran}

\ifCLASSINFOpdf
\else
\fi
\usepackage{amsmath}

\usepackage{lineno,hyperref}
\usepackage{xcolor}
\usepackage{graphicx}
\usepackage[ruled,vlined]{algorithm2e}
\usepackage{algpseudocode}

\newcommand{\blue}[1]{\textcolor{black}{#1}}
\begin{document}
%
\title{Neuroscience-Inspired Algorithms for the Predictive Maintenance of Manufacturing Systems}
%
%
%

\author{Arnav~V.~Malawade,~\IEEEmembership{Student~Member,~IEEE,}
Nathan~D.~Costa,~\IEEEmembership{Member,~IEEE,}
Deepan~Muthirayan,~\IEEEmembership{Member,~IEEE,}
Pramod~P.~Khargonekar,~\IEEEmembership{Fellow,~IEEE,}
Mohammad~A.~Al~Faruque,~\IEEEmembership{Senior~Member,~IEEE}%
\thanks{\copyright 2021 IEEE. Personal use of this material is permitted. Permission from IEEE must be obtained for all other  uses, including reprinting/republishing this material for advertising or promotional purposes, collecting new  collected works for resale or redistribution to servers or lists, or reuse of any copyrighted component of this work in other works.}
\thanks{A. Malawade, N. Costa, D. Muthirayan, P. Khargonekar, and M. Al Faruque are with the Department
of Electrical Engineering and Computer Science, University of California - Irvine, Irvine, CA, 92697 USA e-mail: \{malawada, ndcosta, dmuthira, pramod.khargonekar, alfaruqu\}@uci.edu.}
\thanks{Manuscript received June 11, 2020; revised January 4, 2021.}%
}%
%
%

\markboth{IEEE Transactions on Industrial Informatics,~Vol.~X, No.~X, Month~2021}%
{}

%


\maketitle

\begin{abstract}  
If machine failures can be detected preemptively, then maintenance and repairs can be performed more efficiently, reducing production costs. 
Many machine learning techniques for performing early failure detection using vibration data have been proposed; however, these methods are often power and data-hungry, susceptible to noise, and require large amounts of data preprocessing. 
Also, training is usually only performed once before inference, so they do not learn and adapt as the machine ages.
Thus, we propose a method of performing online, real-time anomaly detection for predictive maintenance using Hierarchical Temporal Memory (HTM). 
Inspired by the human neocortex, HTMs learn and adapt continuously and are robust to noise. 
Using the Numenta Anomaly Benchmark, we empirically demonstrate that our approach outperforms state-of-the-art algorithms at preemptively detecting real-world cases of bearing failures and simulated 3D printer failures. Our approach achieves an average score of 64.71, surpassing state-of-the-art deep-learning (49.38) and statistical (61.06) methods.
\end{abstract}

\begin{IEEEkeywords}
Predictive Maintenance, Prognostics, Anomaly Detection, Hierarchical Temporal Memory
\end{IEEEkeywords}

%

\section{Introduction}
\label{sec:introduction}

\IEEEPARstart{P}{redictive} Maintenance (PM) is an emerging new paradigm in manufacturing where symptoms of machine degradation are detected before failures occur. It is a major part of the Industry 4.0 and smart manufacturing vision.
\blue{Using sensor readings}, process parameters, and other operational characteristics, PM can help maximize tool life by reducing the number of unnecessary repairs performed while also reducing the likelihood of unexpected failures \cite{scheffer2004practical}. 
In the United States alone, improper maintenance and the resulting outages cost more than 60 billion dollars per year \cite{mobley2002introduction}. 
Thus, smart data-driven paradigms such as PM have the potential to \blue{reduce industrial production costs significantly.}

Recently, many statistical, machine learning (ML), and deep learning (DL) techniques for PM have been proposed. 
However, these methods are not without their shortcomings: statistical methods require extensive domain knowledge and often do not generalize well to more complex use cases, while DL and ML techniques often require large amounts of training data and are susceptible to increased error as machines age over time.
Furthermore, ML and DL algorithms are highly susceptible to noise, making them insufficiently robust for industrial settings without data \blue{preprocessing}. Due to the high noise level and diversity among industrial systems, PM models that do not require significant \blue{preprocessing} or domain knowledge are considered more practical \cite{fausing2020predictive}.

To overcome these issues, we propose the use of a learning algorithm inspired by neuroscience called \textbf{Hierarchical Temporal Memory (HTM)}, pioneered by Hawkins and Blakeslee \cite{hawkins2007intelligence}. 
Using binary \textit{sparse distributed representations} (SDRs) to represent data and an architecture incorporating feed-forward, lateral, and feedback connections, HTMs emulate the interactions between pyramidal neurons in the neocortex. 
HTMs are {\it online learning} algorithms that require less application-specific tuning, are robust to noise, 
and adapt to variations in the data as they continuously learn.
\blue{In practice, this means HTMs can efficiently learn from a single training pass over small training datasets with little to no hyperparameter tuning. These characteristics also enable HTMs to learn in near real-time.}
For these reasons, they are suitable for \blue{practical} applications such as detecting early symptoms of failure in manufacturing equipment. In this work, we demonstrate the effectiveness of an HTM-based anomaly detection methodology at detecting these symptoms in roller-element bearings and 3D printers.

\subsection{Related Work}
\label{subsec:relatedwork}
We focus on the specific task of PM on roller-element bearings due to their broad application and utility in manufacturing. We also evaluate Additive Manufacturing (AM) as it is a modern technique that presents unique challenges due to the dynamics of 3D printers. Here, we briefly discuss works related to PM for roller bearings and additive manufacturing.

Many PM methods use statistical models due to their simplicity and explainability. These approaches rely on extracted time and frequency domain features. 
For example, the energy entropy mean and root mean squared (RMS) values of wavelets were used to diagnose ball bearing faults in \cite{seryasat2010multi}.
In another example, the spectral kurtosis (SK) of vibration and current signals was used to detect and classify the surface roughness of ball bearings in \cite{immovilli2009detection}. 
Using a particle filter method, Zhang et al. performed fault detection on bearings similar to those found in helicopter oil cooler fans \cite{zhang2010probabilistic}.

In addition to statistical methods, ML techniques have been applied to a wide array of industrial prognosis tasks.
One such method: AutoRegressive Integrated Moving Average (ARIMA), is one of the most popular techniques for time-series forecasting and was used to predict failures and identify quality defects in a slitting machine in \cite{kanawaday2017machine}. 
In another approach, Tobon-Mejia et al. used Mixture of Gaussians HMMs and Wavelet Packet Decomposition to estimate the Remaining Useful Life (RUL) of roller-element bearings \cite{tobon2012data}.

DL methods such as Long Short-Term Memory (LSTM) Networks and Convolutional Neural Networks (CNNs) have also been used extensively for PM. In one example, Feng et al. used an LSTM for detecting anomalies in industrial control systems \cite{feng2017multi}. 
Additionally, an RNN-LSTM was used to perform PM on an air booster compressor motor used in oil and gas equipment in \cite{abbasi2019predictive}.

Due to the increased complexity and relatively late adoption of AM systems, PM techniques for AM have not been studied in great detail. Proposed approaches often draw from research in related applications, such as PM for bearings. For example, Yoon et al. evaluated the feasibility of AM equipment fault diagnosis using a piezoelectric strain sensor and an acoustic sensor. In this work, features such as RMS value, kurtosis, skewness, and crest factor were used to detect faults \cite{yoon2014phm}. Deep learning has also been used for AM anomaly detection, such as in \cite{yen2019application} where a neural network was used to classify faults in 3D printer vibration data.

\blue{Despite the proliferation of statistical, ML, and DL approaches to PM for manufacturing, to the best of our knowledge, no HTM-based solutions have been proposed.} 
However, the structural and temporal properties of HTM algorithms allow them to excel at cross-domain tasks that \blue{apply} to manufacturing, such as anomaly detection \cite{ahmad2017unsupervised}.
Since the core objective of PM in manufacturing is detecting early symptoms of part failure, HTMs are a natural candidate for this task.
HTMs were shown to match or surpass neural networks at detecting and classifying foreign materials on a conveyor belt in a cigarette manufacturing plant \cite{rodriguez2012raw}. 
\blue{HTMs have also proven effective at detecting anomalies in crowd movements \cite{bamaqa2020anomaly}, traffic patterns \cite{almehmadi2020htm}, human vital signs \cite{bastaki2019application}, electrical grids \cite{barua2020hierarchical}, and computer hardware \cite{faezi2021brain}}.

\subsection{Research Challenges}
Overall, PM for manufacturing presents the following key research challenges:

\begin{enumerate}
    \item \textit{Identifying time-series anomalies in near real-time despite ambient noise.}
    \item \textit{Learning efficiently from small training datasets to improve applicability to practical use cases.}
    \item \textit{Developing a solution that can be generalized to many heterogeneous manufacturing systems without requiring extensive domain-specific tuning.}
    \item \textit{Adapting to changes in data statistics (i.e., machine aging).}
\end{enumerate}

Despite the successes achieved by existing methods in the aforementioned applications, industrial manufacturing systems are diverse and complex, making it difficult to find solutions that generalize across applications. Consequently, PM systems require specialization, which necessitates specialized knowledge and cross-domain skills.
This is especially true in the case of bearing-failure prognosis, as bearing design and lifetime management lies squarely in the mechanical and materials engineering domains. 

It is difficult for any single technique \blue{to address all these research challenges effectively}. For example, statistical methods such as thresholding based on kurtosis or spectral analysis are highly efficient and real-time capable but require explicitly defined health indicators and thresholds, which are machine- and application-specific. 
Also, stationary methods including RMS, kurtosis, and crest factor are only effective for stationary signals (signals with time-invariant statistical properties), but bearing vibration signals are generally cyclostationary (statistical properties vary cyclically) or non-stationary (statistical properties change depending on speed and load conditions)\cite{yan2008feature}.
Spectral kurtosis is applicable to non-stationary and non-periodic signals but is sensitive to noise and outliers \cite{wang2017prognostics}. 

Classical ML algorithms such as AR Models, Support Vector Machines, Hidden Markov Models (HMM), Random Forests, and k-Nearest Neighbors have been demonstrated for PM in existing work, but require the extraction of explicit health indicators (features) from data \cite{wang2018deep}. 
\blue{These algorithms also} require application-specific hyperparameter tuning, data preprocessing as they have poor noise robustness \cite{fausing2020predictive}, and regular updates of model settings as they do not adapt to account for machine aging \cite{wang2018deep}.
Moreover, both HMM and AR methods are ineffective on non-stationary signals \cite{yan2008feature}. 

In DL algorithms such as neural networks and LSTMs, health indicators can be learned implicitly by the network.
However, a network trained for one machine cannot generalize to a new machine without retraining with a large amount of data \blue{for hundreds or thousands of epochs}. Larger models may be able to generalize better, \blue{but} the complexity of training and optimizing these models increases drastically with size \cite{wang2018deep}. \blue{This domain-specific training and tuning process can be expensive, time-consuming, and impractical for real-world use cases.}
Like the ML methods, DL algorithms also have poor noise robustness \cite{kordos2016reducing} and require high-quality data, or else performance can suffer significantly \cite{fausing2020predictive}. To address this, significant preprocessing steps are often needed to generate clean data for these models \cite{fausing2020predictive}.

\blue{As stated in Section \ref{subsec:relatedwork}, HTM-based anomaly detection methods have demonstrated success in several distinct fields. However, to the best of our knowledge, no prior work has comprehensively explored HTM's ability to model vibration data or demonstrated its practical value for PM.}
Overall, all of these existing methods fall short of addressing one or more research challenges.

\subsection{Our Novel Contributions}
To address these key research challenges and improve on the PM performance demonstrated by previous works, our paper presents the following contributions:

\begin{enumerate}
    \item \textit{We demonstrate the ability of HTM-based anomaly detectors to detect early symptoms of bearing failure in several months' worth of real-world vibration data. We show that HTM's can efficiently learn with only a single training pass.}
    
    \item \textit{We demonstrate the ability of HTMs to generalize across applications without much fine-tuning and their ability to continuously learn and adapt by evaluating their anomaly detection performance on a second, highly dynamic application: 3D printer vibration data. These characteristics of HTMs make them more practical for real-world use cases.}
    
    \item \textit{We compare the performance of HTM anomaly detection methods against state-of-the-art anomaly detection techniques and traditional machine prognosis methods such as condition-based maintenance. Specifically, we evaluate each algorithm's anomaly detection accuracy and robustness to noise.}
    
    \item \textit{We demonstrate the efficiency and real-time capability of HTM-based prognosis by comparing its execution time with that of the other techniques.}
\end{enumerate}

\section{Background Theory}
\label{sec:background}

\begin{figure*}[!ht]
    \centering
    \includegraphics[trim=57 243 107 113, clip, width=0.9\textwidth]{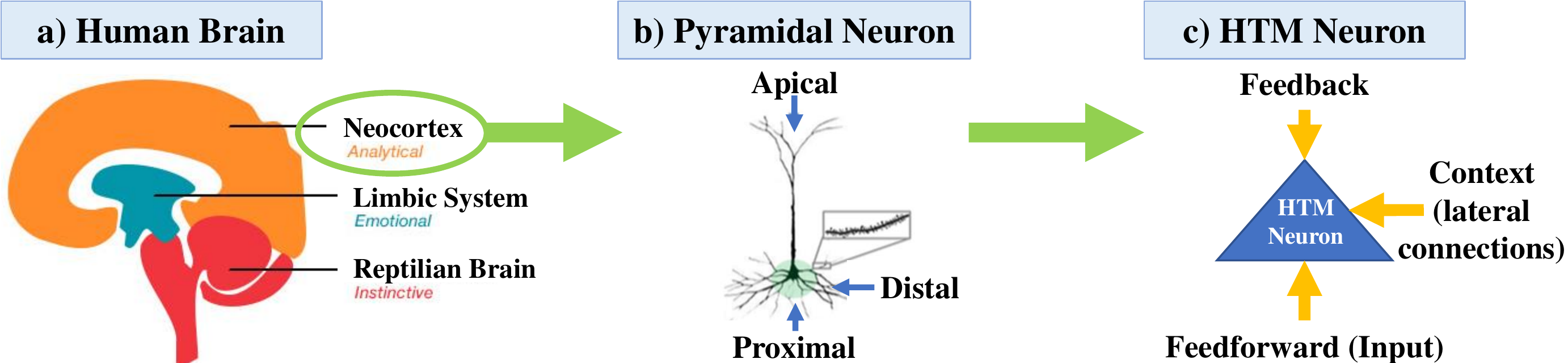}
    \caption{How Neocortical structures are modeled by Hierarchical Temporal Memory. The neocortex is composed of a large number of interconnected pyramidal neurons, each with proximal (feed-forward), apical (feedback), and distal (lateral) dendrites to connect to other neurons. These relations are modeled in HTM neurons as feed-forward, feedback, and lateral connections.}
    \label{fig:brain_to_htm}
    \vspace{-1mm}
\end{figure*}

\subsection{Hierarchical Temporal Memory}
\label{subsec:htm}

Hierarchical temporal memory is a sequence learning framework modeled after the structure of the neocortex in the human brain \cite{hawkins2007intelligence}. 

The basic unit of HTM is a neuron modeled after those present in the neocortex (Fig.~\ref{fig:brain_to_htm}(b)). These neurons are stacked on top of one another to form a column like the `cortical column' of the neocortex. The final HTM is a composition of many such columns. A single HTM neuron (Fig.~\ref{fig:brain_to_htm}(c)), is connected to two types of segments: (i) proximal segments (aggregation of feed-forward connections from the input) and (ii) distal segments (aggregation of lateral connections from neurons of the other columns). Each HTM neuron can be in three states: (i) inactive (the default state), (ii) predictive, and (iii) active. The predictive state of a neuron is determined by the activity of the distal segments, which in turn is determined by the activation state of the other neurons. A neuron becomes active at any time only if it was in the predictive state at the previous instant, with an exception that will be described in Section \ref{subsec:htm-methodology}. When the sequences of activations are viewed temporally, it is easy to see that the distal segments provide the temporal context for activation and thus capture the temporal relations. The column structure augments this capability of HTM by enabling them to store multiple such overlapping temporal sequences. Further details on the HTM-based anomaly detection methodology are discussed in Section \ref{subsec:htm-methodology}.

\subsection{PM of Roller-Element Bearings}
\label{subsec:bearings}
Roller-element bearings perform the critical task of reducing friction between rotating parts in machinery. 
Generally, catastrophic bearing failures present warning signs such as anomalous vibrations and/or noise. These anomalies can occur due to environmental factors (moisture or debris entering the bearing) as well as installation errors (misalignment, excessive loads, or poor/improper lubrication) \cite{ISO15243}. 
Recently, sensor-based techniques that leverage vibration and temperature data to monitor bearing health have been proposed. For example, the NASA Bearing Dataset and the Pronostia Bearing Dataset contain vibration and temperature data for several bearings which were run until failure \cite{lee2007bearing,nectoux2012pronostia}. 
In both datasets, anomalies in the vibration and temperature signals increase in size and frequency as the bearings approach failure, showing a strong correlation between the sensors' readings and system state.

\subsection{PM of 3D Printers}
\label{subsec:3dprinters}
3D printing is a manufacturing process where a physical object is constructed from layers of material in an iterative process. Fused Deposition Modeling (FDM) is a standard technique where melted thermoplastic is extruded through a moving print head nozzle to build each layer. To ensure precision, stepper motors control the extrusion rate of the nozzle as well as the X, Y, and Z-axis movement of the print head.
Since the motors, bearings, and belts are moving parts, they are prone to wear and must be regularly maintained to prevent component failures. As shown in \cite{chhetri2017side}, these components leak vibration information that can be used by PM systems. However, this leaked information is non-stationary since 3D printers move on multiple axes and change direction and speed often, presenting a challenge for conventional PM methods.

\section{Methodology}
\label{sec:methodology}

\subsection{Anomaly Detection using Hierarchical Temporal Memory}
\label{subsec:htm-methodology}
\begin{figure*}[!ht]
    \centering
    \includegraphics[trim=410 225 32 141, clip, width=\textwidth]{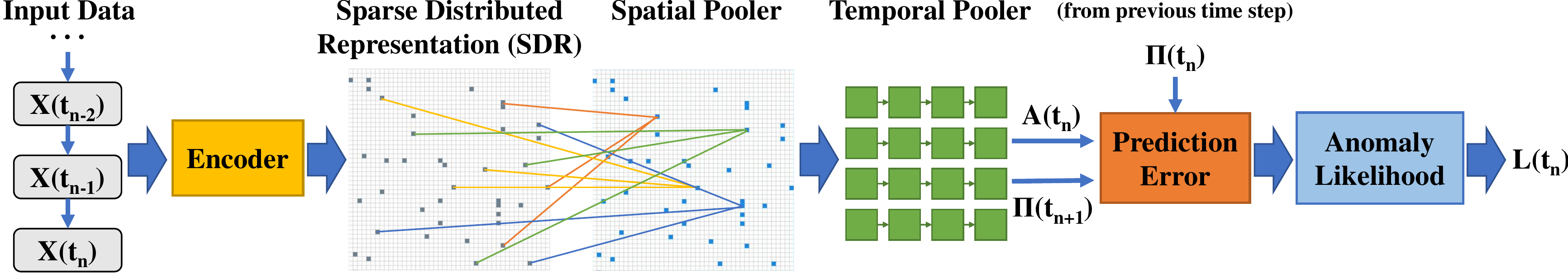}
    \caption{HTM Anomaly Detection Framework. The time-series input $X(t)$ is encoded into a Sparse Distributed Representation (SDR). This information is passed through a spatial pooler and a temporal pooler before outputting a prediction $\Pi(t_{n+1})$ for the next set of column activations. The prediction error between $\Pi(t_n)$ and $A(t_n)$ and the historical distribution of anomaly scores are used to determine the anomaly likelihood $L(t_n)$.}
    \label{fig:htm_methodology}
    \vspace{-4mm}
\end{figure*}

The end-to-end framework for the HTM-based detector is shown in Figure \ref{fig:htm_methodology}. Our methodology for anomaly detection consists of the following steps. First, the time-series vibration data $X(t)$ is taken as input and encoded into a Sparse Distributed Representation (SDR). Next, the SDR is passed through the spatial pooler. \blue{The spatial pooler's output is fed into the temporal pooler, which then outputs a prediction for the next activation $\Pi(t_{n+1})$.} Simultaneously, the prediction from the previous time step $\Pi(t_n)$ is compared with the column activations in the current time step $A(t_n)$ to give a prediction error value: a high error value indicates that this activation was not expected and may be anomalous.
Finally, the anomaly detector uses the \blue{historical} distribution of anomaly scores to calculate the anomaly likelihood $L(t_n)$ for the current data point based on the prediction error value; if $L(t_n)$ exceeds a set threshold, then $X(t_n)$ is flagged as an anomaly. In the following paragraphs, we describe each of these components in detail.

\subsubsection{Encoder}
\blue{The first stage in processing the input data $X(t)$ is the {\em encoder}}. The encoder converts the incoming data point $X(t)$ into a {\em sparse distributed representation} (SDR). This representation is a vector of binary values, \blue{and it is sparse} because only $2\%$ of the bits are activated for any input. This contrasts with deep learning methods that store and learn a dense, distributed representation. Later, we shall describe the advantages of using a sparse representation. We denote the output of the encoder by $x$, a $1 \times n$ vector. 

\subsubsection{Spatial Pooling}
The second stage is {\em spatial pooling}. The spatial pooler identifies spatial relations between different regions of the encoder's output through the proximal connections. Spatial poolers can also be stacked to identify more complex relations. The {\it proximal segment} of each neuron in a column is initialized such that each neuron, where the neurons of the same column share the same proximal segment, is connected to a large fraction of the inputs ($50\%$). The output of this stage is also an SDR representing the columns of the HTM that will be activated in the final output. We denote the {\it spatial pooling} operation mathematically by $I_k(.)$, where the input is the list of columns ordered in decreasing order of their proximal segment values, and k indicates the number of columns to be picked for activation from the top of this list. The number $k$ is typically the top $2\%$, so the output representation is sparse. Let $y_c$ denote the activation of the columns and $P$ denote the proximal connections where $P$ is a binary matrix of size $n \times N$. Then
\begin{equation}
y_c = I_k(xP)
\label{eq:spatialpooling}
\end{equation}

\subsubsection{Prediction}
The next stage is {\it prediction}. The prediction for the next time step is the predictive state of the HTM at the end of the current time step. 
Let the weights of the lateral connections of the $d$th distal segment of the $i$th neuron of $j$th column be $D^d_{i,j}$. 
We note that only those weights of connections which are above a certain threshold are considered to be {\it established} and the rest are set to zero. 
A neuron $(i,j)$ enters the predictive state provided the sum of activations of at least one of the distal segments exceeds a certain threshold, $\theta_d$. Denote the predictive state of a neuron at time $t_n$ by $\pi_{i,j}(t_n)$. We denote the current {\it activation state} of all neurons at time $t_n$ by $A(t_n)$. We denote the total predictive state by the matrix $\Pi(t)$, whose elements are therefore $\pi_{i,j}(t_n)$. Mathematically, $\pi_{i,j}(t_n)$ is given by,
\begin{equation}
\pi_{i,j}(t_n) = \left\lbrace \begin{array}{cc} 1; & \text{if} \ \exists \ d \ \text{s.t.} \ ||D_{i,j}^d \odot A(t_n)||_1 > \theta_d, \\ 0; & \text{otherwise.} \end{array} \right.
\label{eq:predictstate-condn}
\end{equation}
where $\odot$ denotes the element-wise multiplication operation. 

\subsubsection{Temporal Pooling}
The final stage is {\it temporal pooling}. Temporal pooling computes the activation state $A(t_n)$ (an $M\times N$ matrix where M is the number of neurons per mini-column and N is the number of mini-columns in the layer) of the HTM, which is also the output of HTM based on a temporal context. A neuron $i$ is activated provided its column is activated, i.e., $y_c(j) = 1$ and provided it is in the predictive state, i.e., $\pi_{i,j}(t_{n-1}) = 1$. The other neurons in this column are inhibited. If none of the neurons in a column that is active are in the predictive state, then all the neurons of this column are activated. Here, the predictive state $\pi_{i,j}(t_{n-1})$ from the previous time step is the temporal context. This temporal context is updated at the end of this time step as described in the prediction step above. Let $a_{i,j}(t)$ be the $i,j$th element of $A(t_n)$ denoting the activation state of neuron $i$ in column $j$. Then, the temporal pooling operation can be mathematically described as:
\begin{equation}
a_{i,j}(t) = \left\lbrace \begin{array}{cc} 1; & \text{if} \ y_c(j) = 1 \ \text{and} \ \pi_{i,j}(t_{n-1}) = 1, \\  1; & \ y_c(j) = 1 \ \text{and} \ \sum_i \pi_{i,j}(t_{n-1}) = 0, \\ 0; & \text{otherwise.} \end{array} \right. 
\end{equation}

Figure \ref{fig:htm_methodology} shows the different stages of HTM processing in the context of anomaly detection. 
\blue{After activation, the prediction error between the prediction from the previous time step $\Pi(t_n)$ and the current activation state $A(t_n)$ is computed and passed to the anomaly likelihood block, which uses the historical distribution of anomaly scores to determine if $X(t_n)$ is a true anomaly.}

\subsubsection{Learning}
HTMs use a Hebbian-type learning algorithm that reinforces the connection weights of the segments that correctly predict the activation at the next time-step. Each time step, the weights are re-evaluated as follows.
The connection weights of an activated neuron's segments that originated from previously active neurons are increased. The connection weights from neurons that were not active in the previous time-step are decreased. 
Additionally, weights of connections \blue{that are} wrongly predicted are also decreased but at a lesser rate, i.e., forgetting happens at a slower rate than updating. It is this type of learning that allows HTMs to {\it learn continuously and adapt to changes over a long term}. The learning algorithm is discussed in much greater detail in \cite{hawkins2016neurons}.

\subsubsection{On Capacity, Robustness, and Efficiency}
Here, we illustrate why HTMs are efficient and robust to noise. Let us consider an HTM with a large $n$, where $n$ denotes the size of the \blue{encoder's output}, $x$, a binary vector. Denote by $w$ the maximum number of bits that can be one. Typically, $w$ is small relative to $n$. Given this, lets define: $\alpha := w/n$. Here, $\alpha$ is a measure of sparsity and denotes the fraction of the bits that can be active in the SDR of size $n$. An example would be, $n = 2048$ and $w = 4$ and so $\alpha \approx 0.002$. 

The number of possible unique encodings, $N_e$ that can be stored in vector $x$, given $n$ and $w$, is given by,
\begin{equation}
N_e = \left(\begin{array}{c} n \\ w \end{array} \right) = \frac{n!}{w!(n-w)!} 
\end{equation}

For example, if $n = 2048$ and $w = 20$ then $N_e = 10^{47}$. Given $N_e$, the probability that one SDR $x$ will match another SDR $y$, which is randomly picked, is trivially computable:
\begin{equation}
    \mathrm{P}(x = y) = 1/N_e
\end{equation}

Thus, the probability of a false match is\blue{, for all practical purposes, zero}. This shows that SDRs can store and recall reliably an astronomically large number of vectors. Consequently, it follows that HTMs can {\it store and recall reliably an astronomically large number of sequences}.

We can now relax the requirement and say that two SDRs are equivalent if $\theta (< w)$ or more bits match. In this case, the matching is allowed an error of up to $w - \theta$ bits. Denote by $\Omega_x(b)$ the set of sparse vectors (of size $n$ and sparsity $\alpha$) that have an overlap of $b$ bits with $x$. Then, the probability that a false match will be generated, $\mathrm{P}_{fm}$, is given by,
\begin{equation}
\mathrm{P}_{fm} = \frac{\sum_{b \geq \theta}\Omega_x(b)}{N_e}, \ \text{where} \ \Omega_x(b) = \left( \begin{array}{c} w\\b \end{array}\right) \times \left( \begin{array}{c} n-w\\w-b \end{array}\right)\hspace{0.5em}
\end{equation}

Clearly, the probability of a false match has increased by allowing an error of up to $w-\theta$. In the same example as above, if $\theta = 10$, then $w -\theta = 10$, that is an error up to $50 \%$ is allowed. We find that the probability of a false match is still $1/10^{13}$, which for all practical purposes is zero. This is what gives SDRs and thereby HTMs {\it robustness to noise}. 

The sparsity of $x$ allows for sparse computation, \blue{which} makes computations with SDRs very efficient. For a representation $x$ of size $n$ and sparsity $\alpha$, one does not need to store information on all the bits. \blue{Instead,} one can just store the address of the locations of bits of value one. Then, for an operation like matching, one just needs to check the value of the bits of the vector $y$ at its corresponding locations; this is doable almost in constant time.
We can trivially extend this argument to show that the spatial pooling, prediction, and temporal pooling operations described above can also be performed very efficiently in HTMs, thus giving HTMs their {\it computational efficiency}. 
Next, we discuss our experimental setup for demonstrating the performance of the HTM-based anomaly detector.

\subsection{Experimental Setup}
We evaluate our proposed methodology on real-world bearing failure and simulated 3D printer failure datasets. Here, we discuss details about these datasets and the scoring system used for evaluation.

\subsubsection{Bearing Dataset}
\begin{figure*}[!ht]
    \centering
    \includegraphics[trim=312 245 315 160, clip, width=\textwidth]{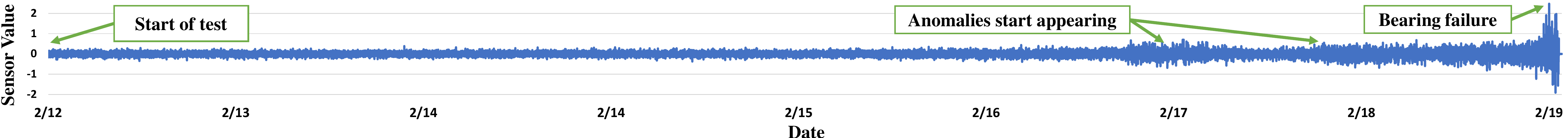}
    \caption{Accelerometer Data from Test 2 of the NASA Dataset \cite{lee2007bearing}. Symptoms of bearing failure can be seen on 2/17 and 2/18 before the bearing's outer race failed on 2/19.}
    \label{fig:bearing-vib-plot}
    \vspace{-4mm}
\end{figure*}

We used the NASA Bearing Dataset and the Pronostia Bearing Dataset \cite{lee2007bearing, nectoux2012pronostia}. The NASA Bearing Dataset contains three tests of bearings run to failure. 
The Pronostia Bearing Dataset contains vibration snapshots recorded with three different radial load and RPM settings.
The accelerometer data for Test 2 of the NASA Dataset is shown in Figure \ref{fig:bearing-vib-plot}. In total, our testing set consists of 40 vibration data files and 191 labeled anomalies.

\subsubsection{3D Printer Dataset}
\begin{figure}[!ht]
    \centering
    \includegraphics[trim=115 40 115 40, clip, width=0.7\linewidth]{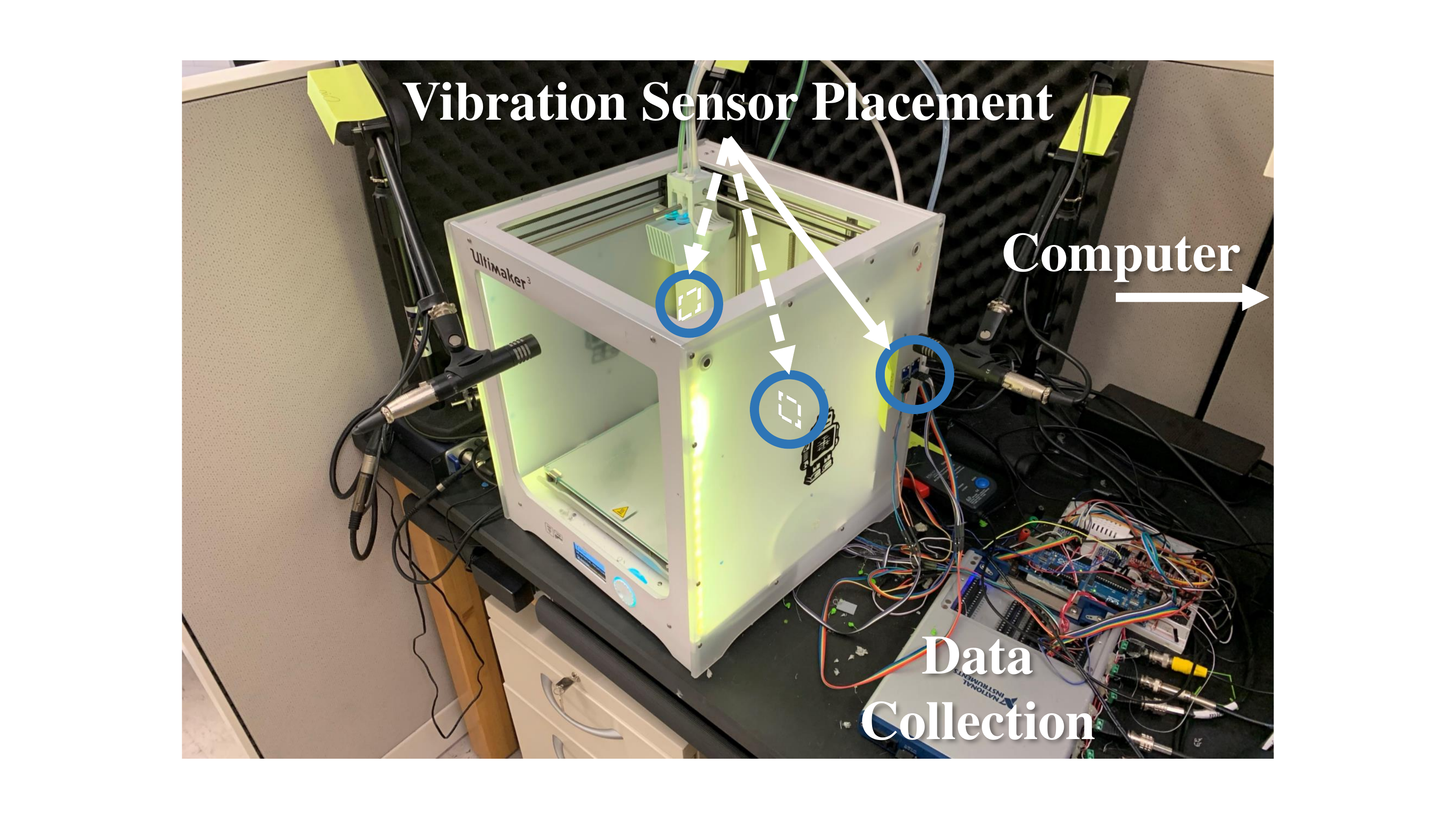}
    \caption{Experimental testbed used to collect vibration data from our 3D printer. Three accelerometers were placed on the printer in total; one sensor was placed directly behind  each of the printer's three stepper motors.}
    \label{fig:3d_testbed}
\end{figure}

Our experimental testbed for collecting vibration data from a 3D printer is shown in Figure \ref{fig:3d_testbed}. The 3D printer uses one stepper motor to control each \blue{movement axis} (X, Y, and Z). We placed one accelerometer directly behind each stepper motor to capture vibration data from prints of various 3D objects. To the best of our knowledge, no publicly available 3D printer component-failure datasets exist\blue{, and} generating real-world failures would risk damaging our equipment. Thus, we instead opted to generate synthetic anomalies in the 3D printer vibration data. 

3D printer vibration signals are inherently non-stationary, meaning that their statistical properties vary with time. However, since printers contain bearings and rotating components with similar dynamics, they share the same time-series and frequency domain features as those correlated with bearing health, such as power spectral density (PSD) \cite{yan2008feature, wang2017prognostics}. For example, in Figure \ref{fig:bearing-vib-plot} it is clear that the overall power of the vibration signal increases as the bearing nears failure. Intuitively, this same phenomenon will occur in a 3D printer as components wear out. Thus, we synthesized anomalies in the 3D printer vibration data by mapping the PSD from our bearing failure data to the 3D printer data. This composition enabled us to simulate the magnitude changes characteristic of bearing and component failures in the 3D printer while preserving the frequency components unique to the 3D printer.

Our PSD mapping algorithm, shown in Algorithm \ref{alg:psd_mapping}, operates on a sliding window over one bearing vibration file and one 3D printer vibration file. For each window $t$, the following steps are performed: First, the Fast Fourier Transform (FFT) $X_b[t]$ of the bearing time-series data $b[n]$ is calculated for a pre-set frequency bin-size. Next, the power in each frequency bin is calculated. Then, we calculate the ratio $C$ between the previous window's power value and the current power value in each bin. This ratio is used to scale the corresponding frequency bin in the FFT of the 3D printer data $FFT(p[t])$, yielding an FFT with synthesized anomalies $X_s[t]$. Finally, the Inverse FFT (IFFT) of $X_s[t]$ is taken and added to the output at location $s[t]$. 

The result after all iterations is a 3D printer vibration signal with synthesized anomalies $s[n]$. Using this mapping algorithm, we produced a simulated 3D printer failure dataset containing 15 test cases and 57 hand-labeled anomalies. 
\begin{algorithm}
\SetAlgoLined
\KwResult{3D printer data with anomalies: $s[n]$}
Initialize bearing and 3D printer data: $b[n]$, $p[n]$\;
Initialize output signal: $s[n] \gets [0, ..., 0]$\;
$t \gets 1$\;
\While{$t < length(p)$}{
$X_b[t] \gets FFT(b[t])$\;
$P_b[t] \gets |X_b[t]|^2$\;
$C \gets \sqrt{P_b[t] / P_b[t-1]}$\;
$X_s[t] \gets FFT(p[t]) \odot C$\;
$s[t] \gets s[t] + IFFT(X_s[t])$\;
$t \gets t + 1$\;
}
\caption{PSD Mapping Algorithm}
\label{alg:psd_mapping}
\end{algorithm}
\vspace{-2mm}

\subsubsection{Anomaly Detectors}
To evaluate the performance of HTMs at PM, we use the following two HTM-based anomaly detectors in our approach with slightly different temporal memory implementations\blue{, which} we denote as HTM \cite{ahmad2017unsupervised} and TM-HTM \cite{numenta2020Feb}. 
\blue{To explore the effectiveness of anomaly likelihood for HTM-based detectors, we evaluated HTM and TM-HTM with three different anomaly likelihood configurations: 
\begin{enumerate}
    \item no anomaly likelihood: the prediction error of the HTM was directly used as the anomaly score. 
    \item historical distribution (HD): the implementation described in Section \ref{subsec:htm-methodology}.
    \item LSTM-based predictor (LP): The HD anomaly likelihood block was replaced with a 2-layer LSTM predictor trained to predict normal HTM prediction error values in order to filter out false positives/noise. The prediction error of the LSTM was used as the final anomaly score.
\end{enumerate}
}

We also evaluated baseline and state-of-the-art anomaly detectors including an RNN-based detector configured to use LSTM cells (denoted as LSTM) \cite{park2018anomaly} (similar to \cite{feng2017multi, abbasi2019predictive}), Windowed Gaussian (based on the tail probability of the distribution over a sliding window), a threshold-based detector (similar to condition-based maintenance and \cite{seryasat2010multi}), EXPoSE \cite{schneider2016expected}, Contextual Anomaly Detector (CAD-OSE) \cite{cadose}, Relative Entropy \cite{wang2011statistical}, Etsy Skyline \cite{skyline}, KNN Conformal Anomaly Detector (KNN-CAD) \cite{burnaev2016conformalized}, Bayesian Changepoint (BC) \cite{adams2007bayesian}, Random (random anomaly score), and Null (constant anomaly score).
All of the listed algorithms except LSTM were exposed to the training data once before testing and updated their models as they were exposed to unseen test data. \blue{LSTM was trained for over 1000 epochs on the training data and was tested with the model settings that resulted in the lowest validation loss. LSTM was tested offline, meaning that it did not update its model weights during testing. The LP anomaly likelihood configuration was also trained in this manner but used the HTM output as its input data instead.}

\subsubsection{Scoring}
To score each algorithm fairly, we rely on the Numenta Anomaly Benchmark (NAB) \cite{ahmad2017unsupervised}. NAB was designed to fairly benchmark anomaly detection algorithms against one another. It contains a built-in anomaly scoring algorithm, normalization, and three threshold optimization settings: standard, low false positives (Low FP), and low false negatives (Low FN). NAB takes in datasets with labeled anomalies and produces \textit{anomaly windows}. These are used to score anomaly detectors on how precisely they can pinpoint anomalies; early/on-time detections are rewarded, and very early/late detections are penalized. 

The NAB scoring function is as follows: given an application profile $A =[A_{TP}, A_{FP}, A_{TN}, A_{FN}]$ specifying the weights for each kind of detection, and the position $y$ of the detection relative to the anomaly window, the scoring function for each detection is:
\begin{equation}
     \sigma^A(y)=(A_{TP} - A_{FP})\left(\frac{1}{(1+e^{5y})} - 1\right)
\end{equation}
\vspace{-3mm}

\begin{figure}[!ht]
    \centering
    \includegraphics[trim=240 190 240 155, clip, width=0.85\linewidth]{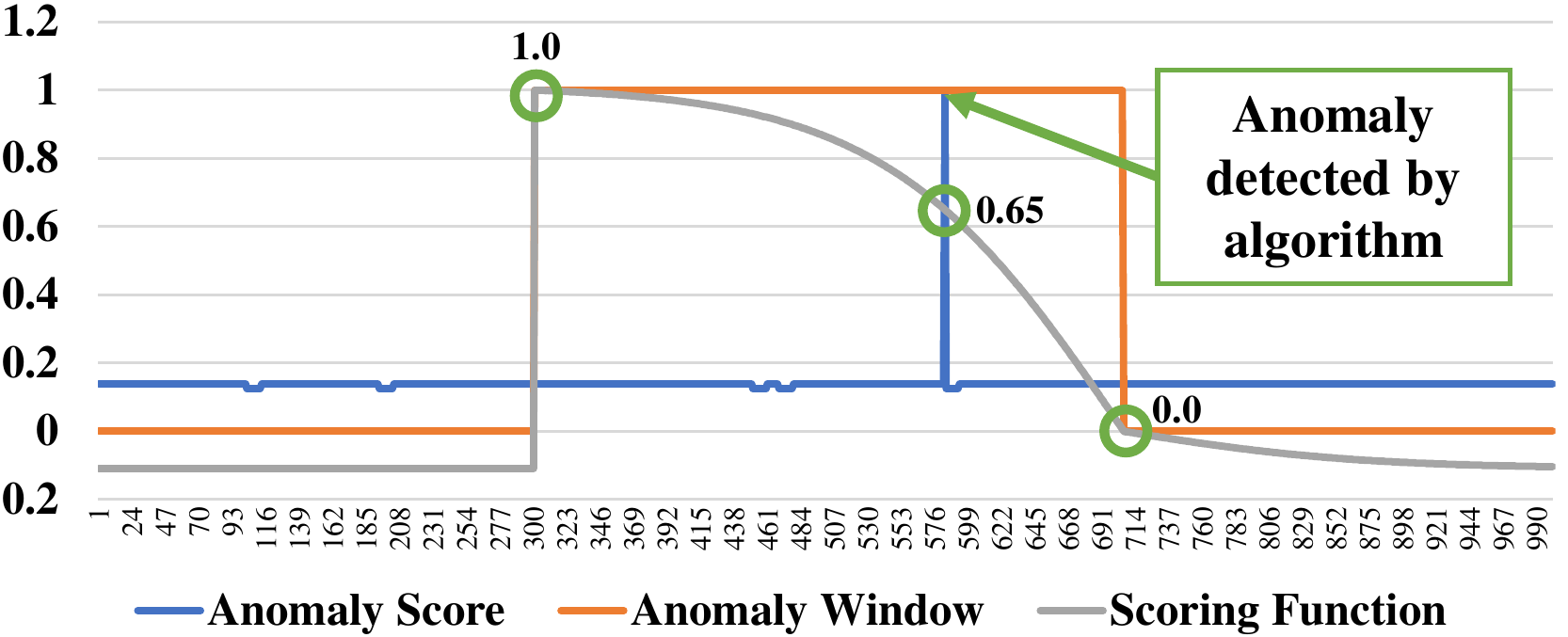}
    \caption{NAB Scoring Functionality: Detection scores are assigned according to the scoring function. The anomaly detected in this example is given a score of 0.65.}
    \label{fig:scoringfunction}
    \vspace{-5mm}
\end{figure}

These scores are summed up for all the detections in a file; the following weighted penalty is deducted for every missed detection ($f_d$): $A_{FN}f_d$. The summed score is then normalized to a 0-100 scale where 0 represents equivalent (or worse) performance to the Null detector\blue{, and} 100 represents a perfect anomaly detector. An example of the scoring functionality is shown in Figure \ref{fig:scoringfunction}. To provide ground-truth values of anomaly locations in the dataset, we followed the NAB official anomaly labeling guide and manually labeled anomalies in each dataset. The first 15\% of each vibration data file was used for training \blue{with the remaining 85\%  used for testing and scoring}. 

\section{Results}
\label{sec:results}

\subsection{Roller Bearing Anomaly Detection}
Table \ref{tab:nab_results} shows the NAB results for the selected algorithms on the labeled bearing failure dataset as well as the total running time of each algorithm. The runtime was recorded over the complete dataset using a PC with an Intel Core i7-7700k processor.
As shown in Table \ref{tab:nab_results}, \blue{TM-HTM+HD achieved the highest anomaly detection score for the Standard and Low FN profiles while HTM+LP achieved the highest score for the Low FP profile. TM-HTM+HD scored \textbf{67.05}, \textbf{73.33}, and \textbf{56.57} for the Standard, Low FN, and Low FP profiles, respectively. The approach that scored closest to HTM was Windowed Gaussian, which achieved scores of \textbf{64.70}, \textbf{70.50}, and \textbf{57.35} for the same profiles, respectively.
HTM and HTM+LP performed better than TM-HTM TM-HTM+LP, indicating that TM-HTM's implementation only works well with the HD anomaly likelihood block. 
}

\blue{As expected, the statistical methods (Windowed Gaussian, Threshold-Based, Relative Entropy) processed the dataset faster than the DL, ML, and HTM-based methods, albeit with lower performance. The HTMs using HD were 1.41x slower than the HTMs with no anomaly likelihood and 3.76x faster than the HTMs using LP on average. TM-HTM+HD processed the dataset \textbf{8.3x faster} than LSTM.}

\begin{table}[!ht]
\centering
\begin{tabular}{p{78pt} p{25pt} p{26pt} p{26pt} p{36pt}}
\hline
Anomaly Detector & \multicolumn{3}{c}{Scoring Profile} & Runtime~(s)\\ \cline{2-4} & Standard & Low~FN & Low~FP \\ \hline
\blue{\textbf{TM-HTM+HD (Ours)}}& \textbf{67.05} & \textbf{73.33} & 56.57 & 4728\\ 
\blue{HTM+HD (Ours)} & 66.38 & 71.93 & 55.33 & 5792\\ 
Windowed Gaussian & 64.70 & 70.50 & 57.35 & 336\\ 
\blue{HTM+LP (Ours)} & \blue{64.03} & \blue{69.12} & \blue{\textbf{57.47}} & \blue{21084}\\
\blue{HTM (Ours)} & \blue{59.75} & \blue{66.24} & \blue{47.63} & \blue{4277}\\
\blue{TM-HTM+LP (Ours)} & \blue{54.12} & \blue{61.53} & \blue{43.03} & \blue{18508}\\
\blue{TM-HTM (Ours)} & \blue{54.39} & \blue{63.33} & \blue{32.47} & \blue{3156}\\
Etsy Skyline \cite{skyline}& 47.53 & 51.51 & 43.75 & 742632\\ 
CAD-OSE \cite{cadose}& 46.88 & 52.81 & 40.96 & 3589\\ 
EXPoSE \cite{schneider2016expected}& 41.75 & 44.80 & 36.96 & 5575\\ 
Threshold-Based & 37.75 & 43.75 & 25.21 & 125\\ 
Relative Entropy~\cite{wang2011statistical} & 34.97 & 37.05 & 32.94 & 806\\ 
LSTM \cite{park2018anomaly}& 33.99 & 38.13 & 28.38 & 43698\\
KNN-CAD \cite{burnaev2016conformalized} & 32.31 & 43.06 &  4.69 & 4393\\ 
Random &  3.06 &  9.16 &  0.00 & 233\\
BC \cite{adams2007bayesian}  &  0.00 &  0.00 &  0.00 & 10270\\ 
Null &  0.00 &  0.00 &  0.00 & 235\\ \hline
    \end{tabular}
    \vspace{4pt}
    \caption{Normalized NAB scores for anomaly detection on the bearing failure dataset.}
    \label{tab:nab_results}
    \vspace{-4mm}
\end{table}

\begin{figure}[!ht]
    \centering
    \includegraphics[trim=233 330 238 337, clip, width=.96\linewidth]{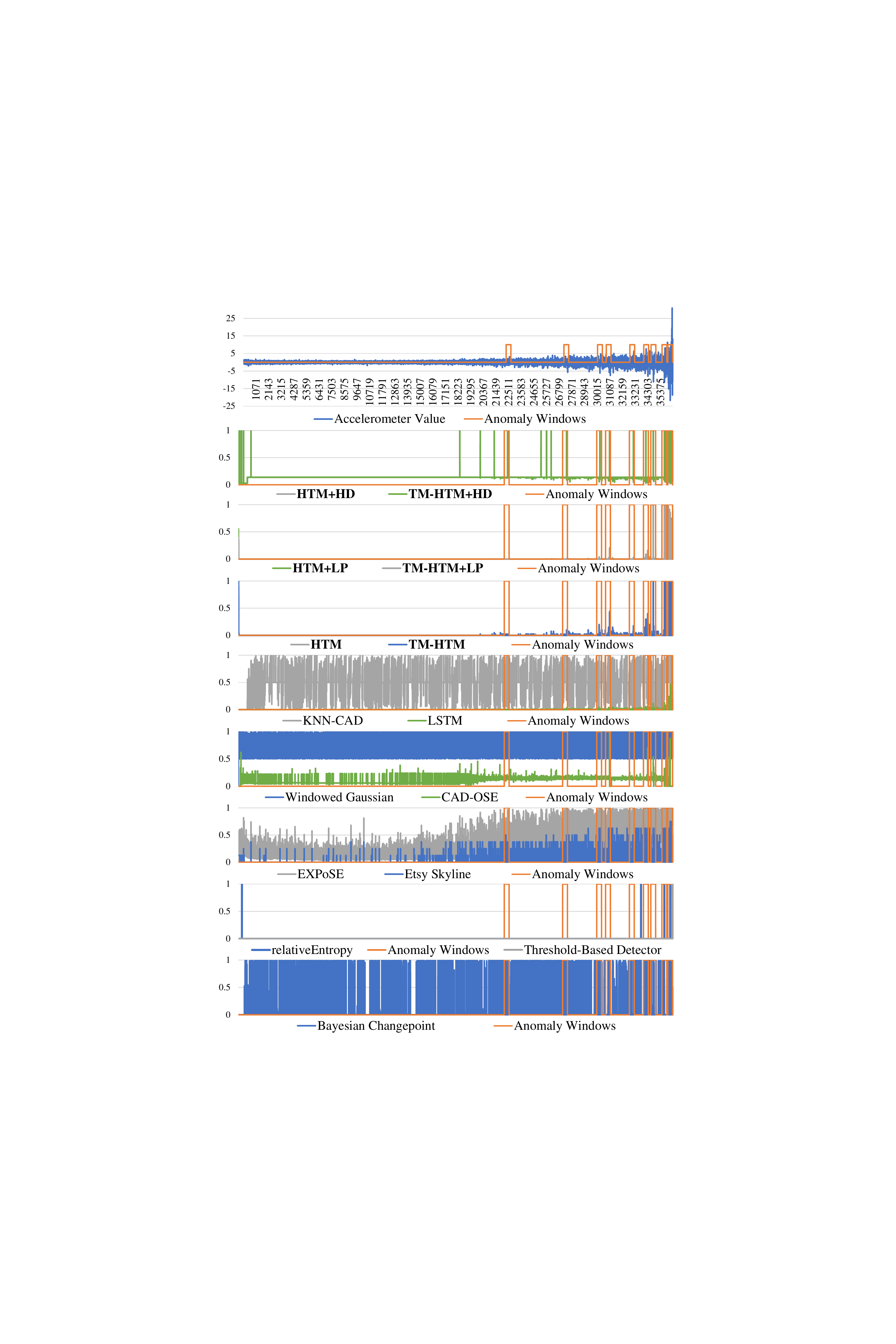}
    \caption{\blue{Anomaly scores for each detector in comparison to the ground truth anomaly windows for Test 1 of the Pronostia Bearing Dataset.}}
    \label{fig:results_plots}
    \vspace{-3mm}
\end{figure}

To evaluate the qualitative performance of each anomaly detector, we plotted the anomaly scores over time for each detector for Test 1 of the Pronostia Bearing dataset and compared them to the labeled ground truth anomaly windows in Figure \ref{fig:results_plots}.

\subsection{3D-Printer Anomaly Detection}
Table \ref{tab:nab_results_3d} shows our experimental results for the 3D printer dataset. \blue{HTM+HD} achieved the highest score on the Low FN profile while LSTM achieved the highest score on the Standard and Low FP profiles. \blue{HTM+HD} achieved scores of \textbf{63.03}, \textbf{73.18}, and \textbf{42.23} for the Standard, Low FN, and Low FP scoring profiles, respectively. LSTM scored \textbf{64.76}, \textbf{71.43}, and \textbf{51.34} at the same profiles, respectively. 
\blue{On both applications the HTM, TM-HTM, and TM-HTM+LP detectors performed worse than the HTM+HD, HTM+LP, and TM-HTM+HD detectors. Overall, the use of HD anomaly likelihood yielded the best HTM performance across applications.}
\blue{Each algorithm's execution time is consistent with the results shown in Table \ref{tab:nab_results}.}

\begin{table}[!ht]
\centering
\begin{tabular}{p{78pt} p{25pt} p{26pt} p{26pt} p{36pt}}
\hline
Anomaly Detector & \multicolumn{3}{c}{Scoring Profile} & Runtime~(s)\\ \cline{2-4}
 & Standard & Low~FN & Low~FP \\ \hline
\blue{\textbf{HTM+HD (Ours)}}& 63.03 & \textbf{73.18} & 42.23 & 5813\\ 
LSTM \cite{park2018anomaly}& \textbf{64.76} & 71.43 & \textbf{51.34} & 16414\\
\blue{TM-HTM+HD (Ours)} & 58.01 & 68.13 & 44.14 & 3364\\
Windowed Gaussian   & 57.42 & 67.99 & 49.31 & 189\\ 

\blue{HTM+LP (Ours)} & \blue{54.56} & \blue{65.45} & \blue{36.76} & \blue{10062}\\
CAD-OSE \cite{cadose} & 54.69 & 63.27 & 34.72 & 1573\\ 
KNN-CAD \cite{burnaev2016conformalized} & 53.76 & 65.55 & 23.48 & 3375\\ 
\blue{HTM (Ours)} & \blue{49.21} & \blue{62.25} & \blue{31.16} & \blue{2713}\\
\blue{TM-HTM+LP (Ours)} & \blue{45.31} & \blue{57.90} & \blue{21.93} & \blue{7456}\\
\blue{TM-HTM (Ours)} & \blue{41.43} & \blue{54.43} & \blue{13.86} & \blue{1724}\\
Relative Entropy \cite{wang2011statistical} & 41.78 & 52.49 & 15.96 & 668\\ 
Etsy Skyline \cite{skyline} & 39.16 & 47.84 & 19.91 & 146165\\ 
BC \cite{adams2007bayesian} & 21.33 & 23.64 & 17.55 & 1946\\ 
Random& 11.34 & 24.75 & 3.74 & 49\\ 
EXPoSE \cite{schneider2016expected}  & 6.04 & 6.20 & 6.04 & 7136\\ 
Threshold-Based  & 0.00 & 0.00 & 0.00 & 54\\ 
Null & 0.00 & 0.00 & 0.00 & 49\\ 
    \hline
    \end{tabular}
    \vspace{4pt}
    \caption{Normalized NAB scores for anomaly detection on the 3D printer dataset.}
    \label{tab:nab_results_3d}
    \vspace{-8mm}
\end{table}

\section{Discussion}
\label{sec:discussion}

\subsection{Overall Performance and Adaptability}
Interestingly, algorithms that performed well on the bearing dataset\blue{, such} as EXPoSE and Etsy Skyline performed worse on the 3D printer dataset. Additionally, algorithms that performed worse on the bearing dataset\blue{, such} as LSTM and BC performed much better on the 3D printer dataset. 
\blue{Our HTM-based methodology using HD anomaly likelihood achieved consistently high performance on both applications without any hyperparameter tuning, demonstrating that this configuration can generalize and adapt to different applications without domain-specific tuning. This result also suggests that HTMs significantly benefit from the inclusion of an HD anomaly likelihood block.}

\blue{Also, HTM+LP was the best performing model on the Low FP profile for the bearing dataset. However, this performance was not replicated in the 3D printer dataset. Similarly, LSTM beat HTM on the Standard and Low FP profile for the 3D printer dataset while performing worse than HTM on the bearing dataset. Hence, our results suggest that LSTMs are highly data-dependent and need to be re-tuned for every machine and/or application. Thus, the LSTM approach is time-consuming, expensive, and impractical for real-world applications.}

The benefits of HTM's continuous learning capability are clearly shown in Figure \ref{fig:results_plots}: after identifying earlier anomalies, \blue{ the HTM-based approaches learn the new baseline for the signal and can pinpoint the future anomalies despite higher signal amplitudes. CAD-OSE also appears to learn continuously, but not as well as the HTMs.}




\subsection{Real-Time Detection Capability}
In addition to detection accuracy and precision, an optimal PM system should be able to detect failure symptoms in real-time to allow adequate time for repairs to be scheduled and performed. 

However, part failures are infrequent and generally present progressive symptoms before failure, so a hard real-time requirement for processing raw sensor data may unnecessarily limit the complexity (and subsequently the performance) of anomaly detection methods. Thus, we evaluate the anomaly detectors in the context of "soft real-time," where we determine if each detector can process a subsampled data segment before the next subsampled data segment arrives. 
For example, 1 second of data can be recorded each minute as a data segment to reduce data size while still ensuring that a wide range of vibration frequencies are captured at frequent intervals.

\blue{Both HTM+HD and TM-HTM+HD were able to process the complete bearing failure dataset in under 100 minutes;} 
since the bearing dataset contains several months' worth of vibration data and minimal data preprocessing was performed (subsampling and timestamping), this demonstrates that HTMs can accurately detect failure symptoms in real-time, meaning that machine operators can be notified of degradation promptly.
Other complex algorithms such as CAD-OSE, KNN-CAD, and EXPoSE had execution time on the same order of magnitude as HTMs and are thus also capable of real-time anomaly detection. \blue{Although HTM+LP, TM-HTM+LP, and LSTM took longer to process the dataset than HTM+HD and TM-HTM+HD, they can still be considered real-time due to the aforementioned dataset characteristics. However, the significant training time associated with the LSTM (over 12 hours on our hardware platform) and the need for application-specific hyperparameter tuning put LSTM at a disadvantage in terms of applicability to practical use cases.}


\subsection{Tunability, and Robustness to Noise}
\label{subsec:tunability}


Figure \ref{fig:results_plots} clearly shows HTM's ability to pinpoint anomalies while remaining robust to noise in the input.
\blue{This is likely due to HTMs use of sparse encodings, making it unlikely that bit errors in the input due to noise will affect the bits corresponding to the input pattern, making them robust to noise.
From the figure, it is also clear that the HTM implementations using anomaly likelihood blocks were more robust to noise outside of the anomaly windows than the HTM or TM-HTM alone. This is likely because the anomaly likelihood components filter out smaller detections to isolate only the most plausible anomalies. The HTM+HD and TM-HTM+HD detected anomalies earlier than the other configurations, albeit with slightly more false positives.
The outputs of the different HTMs starkly contrast with the highly variable anomaly score outputs of Windowed Gaussian, EXPoSE, KNN-CAD, and BC, among others.} These detectors record high anomaly scores even when there is relatively low noise in the input, meaning that they will likely suffer from false positives at higher noise levels. 

A detector's threshold can be tuned to account for higher noise levels; however, for detectors such as Windowed Gaussian\blue{, which} used the maximum detection threshold of 1.0, the threshold cannot be increased further to reduce its sensitivity. 
\blue{In contrast, TM-HTM+HD used a threshold of 0.5497 on the Standard profile. 
Thus, although Windowed Gaussian outperformed TM-HTM+HD on the Low FP scoring profile, it lacks tunability and will likely perform much worse than this HTM configuration in more noisy environments. }


LSTM appears to have good robustness to noise\blue{, as} shown in Figure \ref{fig:results_plots}. However, it is clear from the figure that it missed some of the earlier anomaly windows completely. In the context of PM, this can mean that an observer will only be warned of degradation later and will not have much time to organize repairs. 
Overall, our methodology demonstrates significant noise-robustness, better tunability, and the ability to detect early anomalies as well as larger, late-stage anomalies. 

\subsection{Limitations and Future Work}
Another related PM problem is Remaining Useful Life (RUL) estimation. In many cases, RUL and anomaly detection go hand in hand as part of a comprehensive PM system. Although we did not evaluate the performance of HTM at RUL estimation, the core architecture of HTM is good at sequence prediction and could likely be used to solve this problem. We leave this for future work.

Another limitation of our work is the use of synthesized 3D printer anomalies instead of real-world examples of 3D printer failures. Due to resource constraints, we opted not to perform these experiments and used synthetic failure data instead. The question of whether HTM's performance on synthetic anomalies translates to real-world PM remains an open research problem.

\subsection{Feasibility}
The idea of predicting machine failures in advance is not brand new; many variants of PM systems have already been implemented in real-world manufacturing applications. However, based on our results, we believe that HTM is a better solution than current state-of-the-art methods.
Our results demonstrate that HTMs are efficient enough to run on consumer-grade processors while learning and adapting continuously. Additionally, HTMs can be easily installed on existing PM systems as they only require time-series sensor inputs, which likely already exist in the system.
\blue{As shown by our results, the industry-standard LSTM requires a significant amount of time for training (over 1000 epochs) as well as application-specific tuning.
In contrast, HTMs do not require any application-specific parameter tuning and are essentially plug-and-play since they only need to be trained with a single pass on normal sensor data}. These characteristics make HTMs an extremely viable, out-of-the-box solution for industrial PM.

\section{Conclusion}
\label{sec:conclusion}
Existing methods for predicting machine failures from sensor data are limited in their practicality due to shortcomings\blue{, including} poor noise resistance, efficiency, and adaptability. \blue{Our experiments demonstrated} that our methodology outperforms state-of-the-art approaches at detecting anomalies in both bearing and 3D printer failure data with minimal to no pre-processing or application-specific tuning.
On the Standard scoring profile, \blue{our methodology using HD anomaly likelihood} achieved an average NAB score of \textbf{64.71}. In comparison, the other top algorithms: LSTM and Windowed Gaussian, achieved average scores of 49.38 and 61.06, respectively. 
Furthermore, our qualitative results show that our methodology is significantly more noise-resistant than the Windowed Gaussian, KNN-CAD, EXPoSE, and BC detectors, which we attribute to the use of SDRs and an anomaly likelihood component.
We also demonstrated that our methodology is real-time capable, with an execution time on the same order of magnitude as state-of-the-art methods.
Consequently, we conclude that HTM-based anomaly detection is a novel, practical solution for a wide range of industrial PM applications.

\section*{Acknowledgment}
This work was partially supported by the National Science Foundation (NSF) under awards CMMI-1739503 and ECCS-1839429 as well as by Graduate Assistance in Areas of National Need (GAANN) under award P200A180052. 
Any opinions, findings, conclusions, or recommendations expressed in this paper are those of the authors and do not necessarily reflect the views of the funding agencies.

\bibliography{IEEEabrv, bibliography}

\ifCLASSOPTIONcaptionsoff
  \newpage
\fi


%

\begin{IEEEbiography}[{\includegraphics[width=1in, height=1.25in, clip, keepaspectratio]{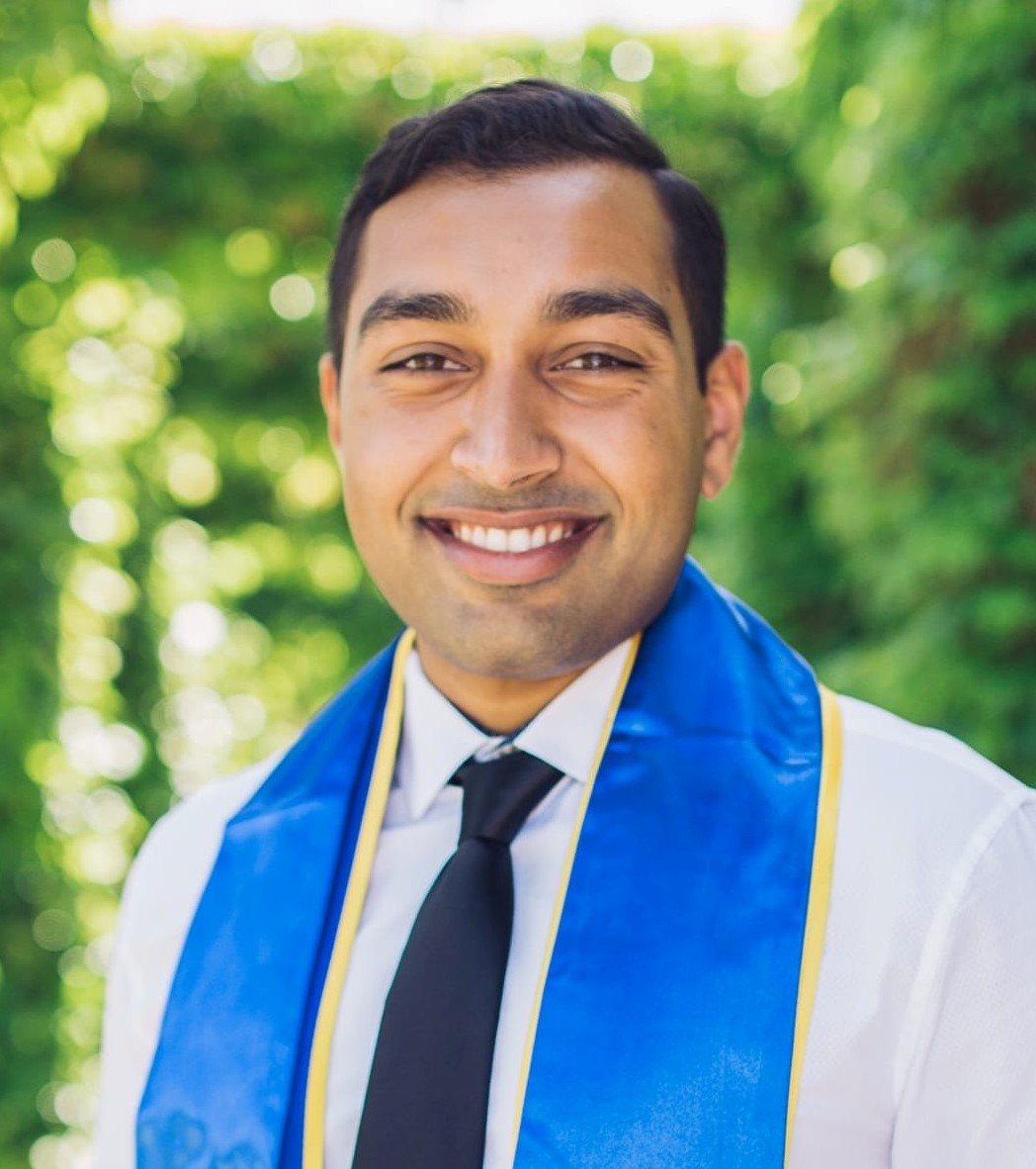}}]{Arnav V. Malawade}
received a B.S. in Computer Science and Engineering from the University of California Irvine (UCI) in 2018. He is currently an M.S./Ph.D. Student studying  Computer Engineering at UCI under the supervision of Professor Mohammad Al Faruque. His research interests include the design and security of cyber-physical systems in connected/autonomous vehicles, manufacturing, IoT, and healthcare.
\end{IEEEbiography}


\begin{IEEEbiography}[{\includegraphics[width=1in, height=1.25in, clip, keepaspectratio]{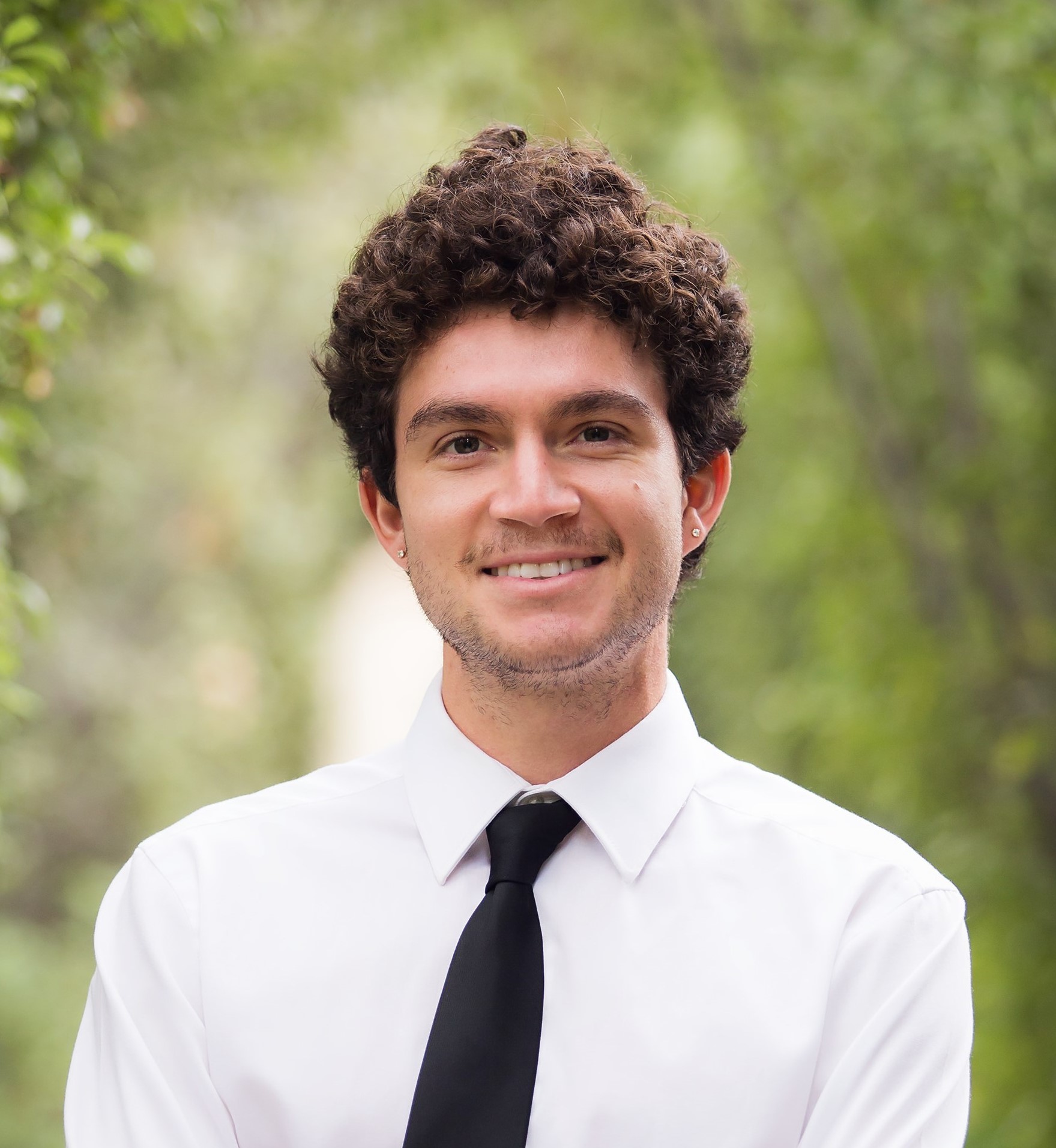}}]{Nathan D. Costa}
received a B.S. in Computer Science and Engineering from the University of California Irvine (UCI) in 2020. He is currently applying to industries relevant to his interests, those being embedded software development and embedded system design. 
\end{IEEEbiography}


\begin{IEEEbiography}[{\includegraphics[width=1in, height=1.25in, clip, keepaspectratio]{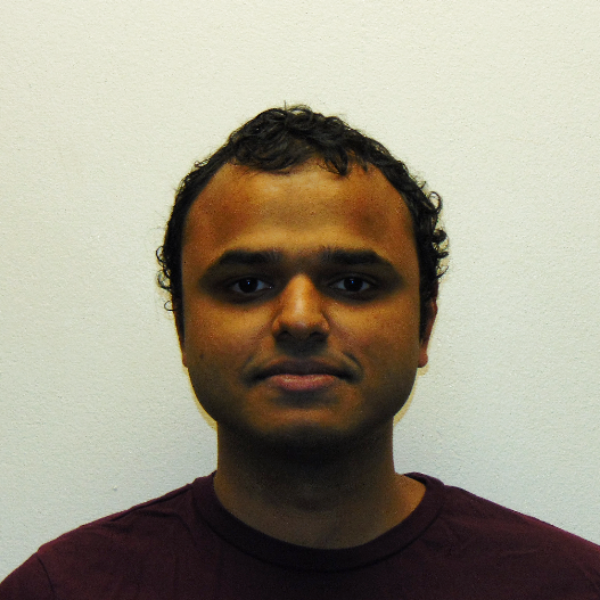}}]{Deepan Muthirayan}
is currently a Post-doctoral Researcher in the department of Electrical Engineering and Computer Science at University of California at Irvine. He obtained his Phd from the University of California at Berkeley (2016) and B.Tech/M.tech degree from the Indian Institute of Technology Madras (2010). His doctoral thesis work focused on market mechanisms for integrating demand flexibility in energy systems. Before his term at UC Irvine he was a post-doctoral associate at Cornell University where his work focused on online scheduling algorithms for managing demand flexibility. His current research interests include control theory, machine learning, topics at the intersection of learning and control, online learning, online algorithms, game theory, and their application to smart systems.
\end{IEEEbiography}


\begin{IEEEbiography}[{\includegraphics[width=1in, height=1.25in, clip, keepaspectratio]{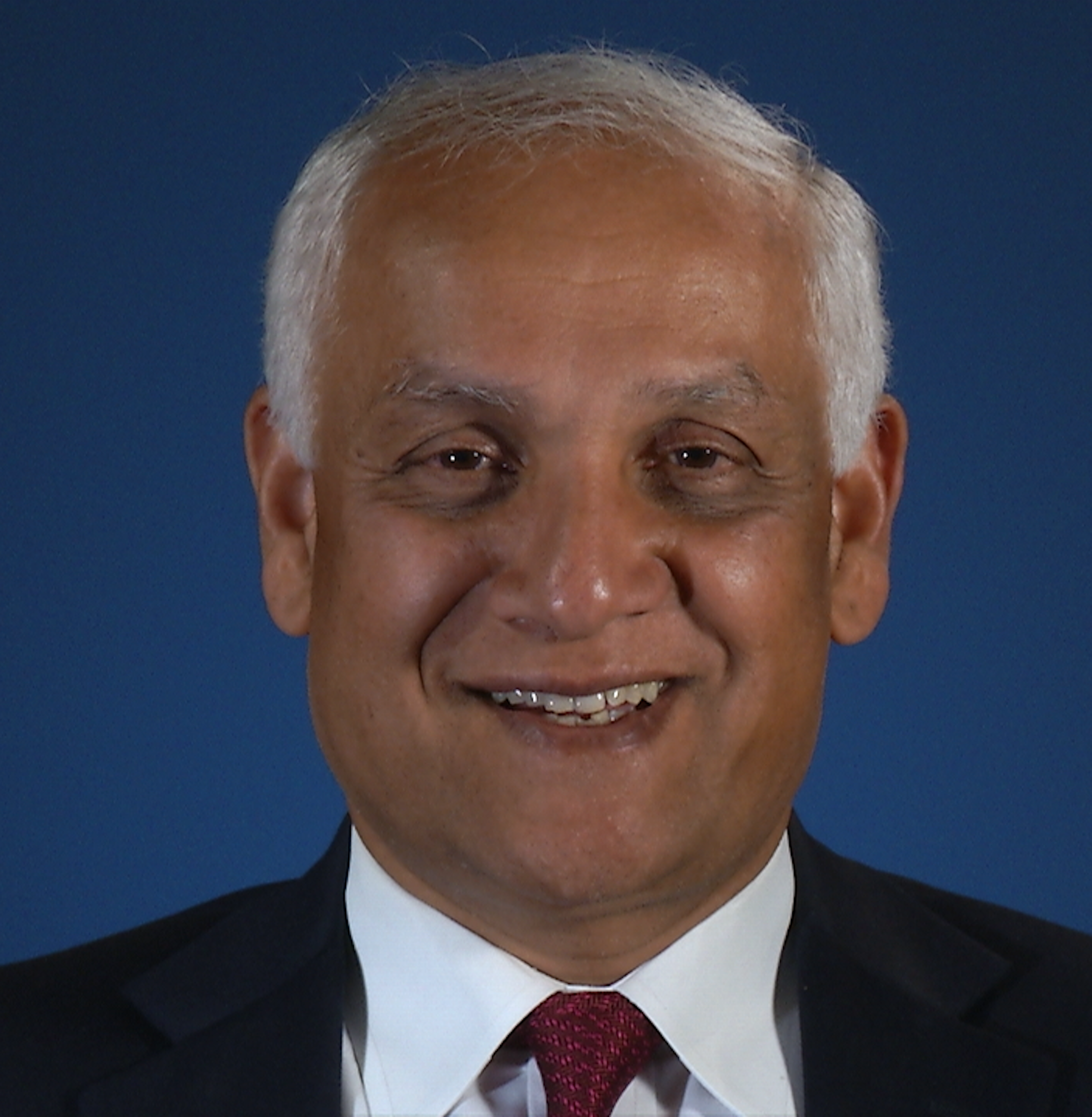}}]{Pramod~P.~Khargonekar}
received B. Tech. Degree in electrical engineering in 1977 from the Indian Institute of Technology, Bombay, India, and M.S. degree in mathematics in 1980 and Ph.D. degree in electrical engineering in 1981 from the University of Florida, respectively. He was Chairman of the Department of Electrical Engineering and Computer Science from 1997 to 2001 and also held the position of Claude E. Shannon Professor of Engineering Science at The University of Michigan.  From 2001 to 2009, he was Dean of the College of Engineering and Eckis Professor of Electrical and Computer Engineering at the University of Florida till 2016. After serving briefly as Deputy Director of Technology at ARPA-E in 2012-13, he was appointed by the National Science Foundation (NSF) to serve as Assistant Director for the Directorate of Engineering (ENG) in March 2013, a position he held till June 2016. Currently, he is Vice Chancellor for Research and Distinguished Professor of Electrical Engineering and Computer Science at the University of California, Irvine. His research and teaching interests are centered on theory and applications of systems and control. He has received numerous honors and awards including IEEE Control Systems Award, IEEE Baker Prize, IEEE CSS Axelby Award, NSF Presidential Young Investigator Award, AACC Eckman Award, and is a Fellow of IEEE, IFAC, and AAAS.
\end{IEEEbiography}


\begin{IEEEbiography}[{\includegraphics[width=1in, height=1.25in, clip, keepaspectratio]{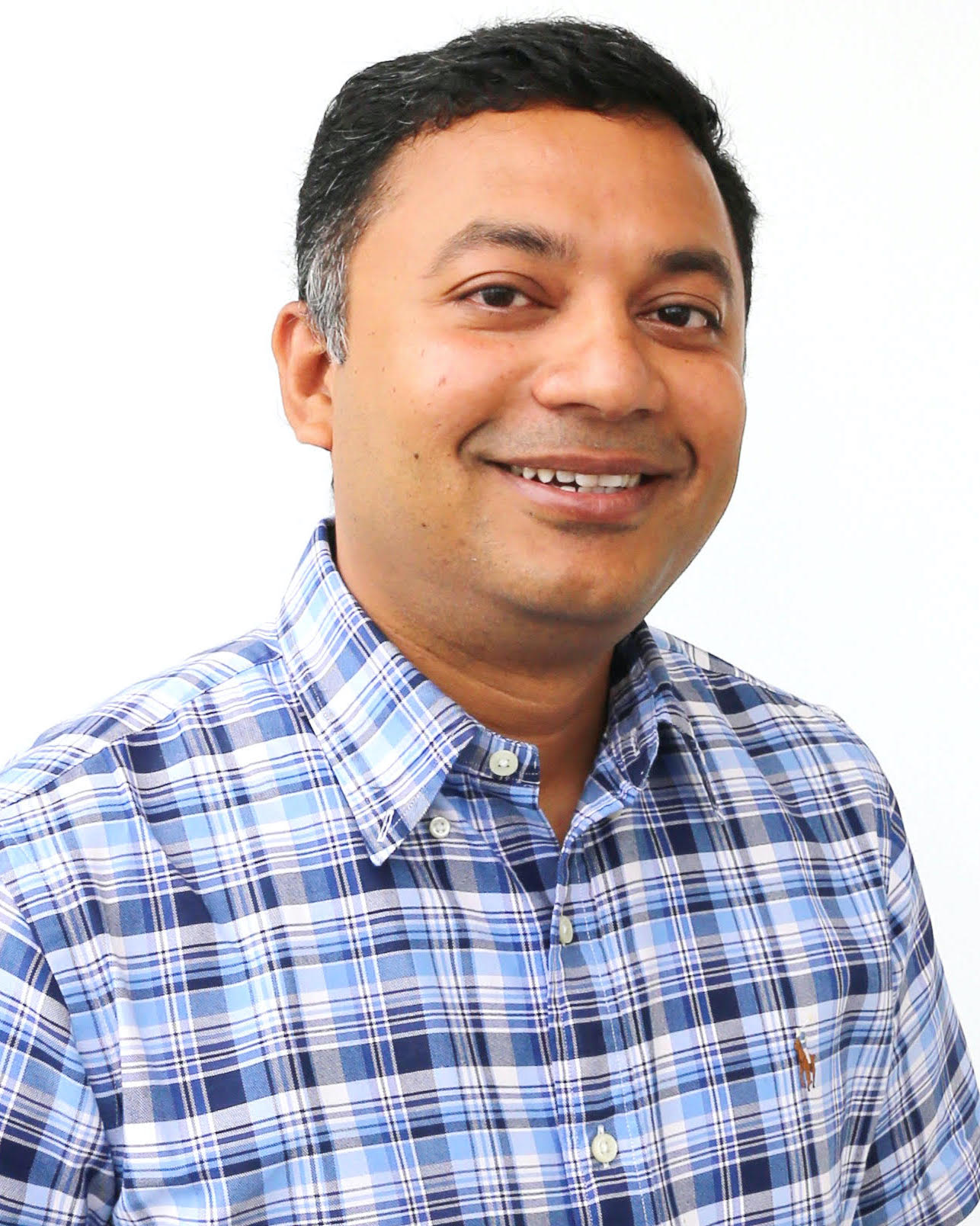}}]{Mohammad A. Al Faruque}
(M’06, SM’15) received his B.Sc. degree in Computer Science and Engineering (CSE) from Bangladesh University of Engineering and Technology (BUET) in 2002, and M.Sc. and Ph.D. degrees in Computer Science from Aachen Technical University and Karlsruhe Institute of Technology, Germany in 2004 and 2009, respectively.
He is currently with the University of California Irvine (UCI) as an Associate Professor and Directing the Embedded and Cyber-Physical Systems Lab. He served as an Emulex Career Development Chair from October 2012 till July 2015. Before, he was with Siemens Corporate Research and Technology in Princeton, NJ as a Research Scientist. His current research is focused on the system-level design of embedded and Cyber-Physical-Systems (CPS) with special interest in low-power design, CPS security, data-driven CPS design, etc.
He is an ACM senior member. He is the author of 2 published books. Besides many other awards, he is the recipient of the School of Engineering Mid-Career Faculty Award for Research 2019, the IEEE Technical Committee on Cyber-Physical Systems Early-Career Award 2018, and the IEEE CEDA Ernest S. Kuh Early Career Award 2016. He is also the recipient of the UCI Academic Senate Distinguished Early-Career Faculty Award for Research 2017 and the School of Engineering Early-Career Faculty Award for Research 2017. Besides 120+ IEEE/ACM publications in the premier journals and conferences, he holds 9 US patents.
\end{IEEEbiography}





\end{document}